\documentclass{article}
\usepackage{multicol}
\usepackage{PRIMEarxiv}

\usepackage[utf8]{inputenc}   
\usepackage[T1]{fontenc}      
\usepackage{url}              
\usepackage{booktabs}         
\usepackage{amsfonts}         
\usepackage{nicefrac}         
\usepackage{fancyhdr}         
\usepackage{graphicx}         
\usepackage{amsmath}
\usepackage{inconsolata}
\usepackage{microtype}        
\usepackage{multirow}         
\usepackage{hyperref}         

\providecommand{\Description}[1]{}

\title{Iterated Integrals as Lexical Fingerprints: Path Signature Embeddings for Lexical Similarity}

\author{
  Raj Shekhar Singh \\
  Indian Institute of Technology, Roorkee \\
  Roorkee, Uttarakhand, India \\
  \texttt{raj\_ss@ece.iitr.ac.in}
  \and
  Pratham Agarwal \\
  Indian Institute of Technology, Roorkee \\
  Roorkee, Uttarakhand, India \\
  \texttt{pratham\_a@ece.iitr.ac.in}
}

\begin{document}
\maketitle

\begin{abstract}
While sentence embeddings are primarily designed for semantic generalization, their latent geometry implicitly encodes surface-level lexical properties, a relationship that remains poorly understood. Shallow pooling methods discard word order, while deep Transformer models obscure these lexical alignments behind computational complexity and uninterpretable spaces. We propose constructing sentence embeddings via the path signature of Word2Vec trajectories, leveraging the iterated integrals of rough path theory to create representations that explicitly capture sequential and lexical structure in a geometrically tractable manner. Through extensive ablations spanning Word2Vec dimensions, signature depths, and PCA compression, we correlate embedding distances against five lexical string metrics: Levenshtein, Jaro-Winkler, word Jaccard, phoneme Levenshtein, and character TF-IDF. We reveal that while average pooling almost entirely fails to capture character-level edit distances, path signatures strongly align with them, with depth 2 signatures optimally balancing character and word-level sensitivities. Furthermore, we demonstrate that the critical lexical variance is concentrated in a highly compact PCA subspace, and unlike pooling, signatures structurally preserve and quantify word order. These findings establish path signatures not as semantic generalizers, but as rigorous mathematical lenses that explicitly encode discrete lexical structures and word order, offering an interpretable alternative to opaque deep learning for lexical tracking.
\end{abstract}
\begin{multicols}{2}
\section{Introduction}
Sentence embeddings map discrete text into continuous vector spaces, serving as foundational components in modern NLP pipelines for tasks such as semantic similarity and information retrieval \cite{cer2018universal}. While the primary objective of these representations is typically to capture high-level semantics, their latent geometry often implicitly encodes surface-level lexical properties. Metrics such as Levenshtein distance \cite{lcvenshtcin1966binary}, character n-gram overlap, and phonetic similarity act as strong baselines for text relatedness and are particularly vital in domains sensitive to morphological variation or typographical errors. However, the exact algebraic mechanisms by which continuous sentence representations absorb and reflect these discrete, string-level metrics remain poorly understood. Investigating this connection is essential for understanding how continuous representations model discrete lexical structures, yet methods explicitly designed to encode and study these lexical and sequential properties remain underexplored.

Existing methods for constructing sentence embeddings occupy two extremes, each bearing significant limitations. On one end, shallow pooling operations, such as averaging or maxing pre-trained word vectors \cite{mikolov2013efficient, church2017word2vec}, are computationally inexpensive but fundamentally permutation invariant. By treating sentences as unordered bags of words, these approaches entirely discard positional information, rendering them incapable of distinguishing sequences with identical words but different syntactic structures. On the opposite end, deep learning approaches, particularly Transformer-based models \cite{vaswani2017attention, devlin2019bert}, achieve state-of-the-art semantic fidelity but incur substantial computational costs. Furthermore, the high-dimensional, non-convex latent spaces learned by these models are largely uninterpretable, obscuring their sensitivity to lexical perturbations and making it difficult to analytically quantify their reliance on surface-level features without exhaustive empirical probing.

In this work, we propose constructing sentence embeddings via the path signature, a mathematical transform from rough path theory \cite{lyons1998differential, Chen1958INTEGRATIONOP} defined by the iterated integrals of a continuous path, to explicitly model the lexical and sequential structure of text. By mapping a sentence to a piecewise linear path through its sequential Word2Vec \cite{mikolov2013distributed} waypoints, the path signature naturally encodes both the lexical identity of individual tokens and their sequential interactions, overcoming the permutation invariance of standard pooling. We conduct a comprehensive empirical study sweeping across Word2Vec architectures, embedding dimensions, and signature depths \cite{reizenstein2018iisignature, kidger2020signatory} and PCA compression to manage the resultant high dimensionality. We evaluate these representations against rigorous lexical metrics. Our findings reveal that signature embeddings function primarily as explicit encoders of lexical and sequential structure rather than abstract semantic representations. While they do not replicate the semantic generalization of deep contextual models, they strongly align with discrete string metrics and structurally preserve word order. Their capacity to capture these lexical properties is highly dependent on signature configuration, with appropriate depth and PCA-based compression substantially improving the fidelity of the lexical encoding.

\section{Related Work}

\subsection{Sentence-Level Embeddings}
Sentence embedding methods range from static aggregations to deeply contextualized representations. Early approaches pool pre-trained word vectors (e.g., Word2Vec \cite{church2017word2vec}), which is efficient but permutation invariant, entirely discarding word order and syntactic structure. Transformer-based models like BERT \cite{devlin2019bert} and Sentence-BERT \cite{reimers2019sentence} resolve this via attention mechanisms and contrastive fine-tuning, achieving state-of-the-art semantic fidelity. However, these highly parameterized models incur substantial computational costs and operate as opaque systems, sacrificing geometric interpretability. This leaves a gap for representations that are sequentially aware yet mathematically tractable.

\subsection{Path Signatures}
The path signature, a construct from rough path theory \cite{lyons1998differential, Chen1958INTEGRATIONOP}, characterizes continuous paths via an infinite collection of iterated integrals:
\begin{equation}
S(X)_{0,1} = \left(1, \int_{0<s<1} dX_s, \int_{0<s_1<s_2<1} dX_{s_1} \otimes dX_{s_2}, \dots \right).
\end{equation}
Uniquely defining the path up to tree-like equivalence, the signature acts as a universal non-linear feature map \cite{chevyrev2025primer, cass2024lecture} that inherently captures sequential order and hierarchical interactions \cite{kiraly2019kernels} without recurrent or attention mechanisms. While highly successful in time-series and handwriting recognition \cite{xie2017learning, yang2022developing} due to parameterization invariance, NLP applications remain nascent \cite{tseriotou2024sig, kormilitzin2020efficient}. The systematic relationship between the geometry of signature text embeddings and discrete lexical metrics, modulated by depth and dimensionality, remains unexplored.

\section{Experiments}
All experiments were executed on a dual NVIDIA T4 GPU configuration. We pretrain Word2Vec \cite{mikolov2013efficient} embeddings on a 100,000-sentence stream of BookCorpus \cite{zhu2015aligning} and evaluate the derived sentence representations on the STS-12 benchmark \cite{agirre2012semeval}. To rigorously quantify how embedding geometries reflect surface-level text properties, we compute distances in the embedding space and correlate them against five distinct lexical metrics: Levenshtein distance \cite{lcvenshtcin1966binary}, Jaro-Winkler distance \cite{winkler1990string}, word Jaccard distance, phoneme Levenshtein distance, and character TF-IDF \cite{salton1988term} cosine distance. Alignment is assessed using three standard correlation coefficients: Pearson, Spearman, and Kendall $\tau$. We benchmark the signature-based approach against the conventional mean pooling of the corresponding Word2Vec embeddings to isolate the specific contribution of the path transform.

We conduct a multifaceted ablation study to isolate the structural contributions of the path signature. We first vary the Word2Vec output dimension across {25, 50, 100, 200, 400} to observe how token-level geometric resolution propagates to the sentence level. We then increment the signature depth from 1 to 3, evaluating the marginal utility of capturing higher-order sequential interactions via iterated integrals. Given the combinatorial explosion of the signature feature space, we apply PCA compression at multiple variance retention thresholds to probe the intrinsic dimensionality of the embeddings and assess the trade-off between dimensionality reduction and lexical alignment retention. Finally, we perform a text positioning study to quantify the sequential sensitivity of the representations. By randomly permuting the tokens within sentences and measuring the resulting normalized inversion disorder, we compute the correlation between embedding distance and positional perturbation, explicitly testing whether path signatures capture word order independently of bag-of-words statistics.

\begin{table*}[t]
\centering
\caption{Pearson correlation of embedding distances against five lexical string metrics across varying Word2Vec dimensions and signature configurations.}
\label{tab:lexical_corr}
\vspace{0.1cm}
\resizebox{\textwidth}{!}{%
\begin{tabular}{llccccc}
\toprule
\textbf{Configuration} & \textbf{Dim} & \textbf{Levenshtein} & \textbf{Jaro-Winkler} & \textbf{Word Jaccard} & \textbf{Phoneme Lev.} & \textbf{Char TF-IDF} \\
\midrule
\multirow{4}{*}{Avg Pooling}
    & 50  & 0.036 & 0.521 & 0.662 & 0.017 & 0.558 \\
    & 100 & 0.023 & 0.540 & 0.681 & 0.003 & 0.560 \\
    & 200 & 0.019 & 0.544 & 0.682 & 0.001 & 0.555 \\
    & 400 & 0.018 & 0.544 & 0.681 & 0.001 & 0.553 \\
\midrule
    Sbert Text2Vec & -  & 0.218 & 0.502 & 0.517 & 0.211 & 0.461 \\
    Qwen Embed & -  & 0.063 & 0.387 & 0.561 & 0.063 & 0.716 \\
    Minilm Embed & - & 0.025 & 0.406 & 0.608 & 0.031 & 0.764 \\
\midrule
\multirow{4}{*}{Sig (depth 1)}
    & 50  & 0.367 & 0.547 & 0.352 & 0.361 & 0.226 \\
    & 100 & 0.367 & 0.567 & 0.357 & 0.359 & 0.218 \\
    & 200 & 0.355 & 0.571 & 0.356 & 0.357 & 0.209 \\
    & 400 & 0.361 & 0.572 & 0.354 & 0.357 & 0.206 \\
\midrule
\multirow{4}{*}{Sig (depth 2)}
    & 50  & 0.351 & 0.609 & 0.690 & 0.354 & 0.529 \\
    & 100 & 0.352 & 0.613 & 0.693 & 0.356 & 0.522 \\
    & 200 & 0.354 & 0.612 & 0.690 & 0.359 & 0.519 \\
    & 400 & 0.354 & 0.607 & 0.690 & 0.359 & 0.513 \\
\midrule
    Sig (depth 3)
    & 50  & 0.388 & 0.614 & 0.699 & 0.393 & 0.528 \\
\midrule
\multirow{4}{*}{Sig (depth 2, PCA 10\%)}
    & 50  & 0.292 & 0.593 & 0.688 & 0.299 & 0.552 \\
    & 100 & 0.281 & 0.607 & 0.695 & 0.289 & 0.554 \\
    & 200 & 0.277 & 0.604 & 0.689 & 0.286 & 0.543 \\
    & 400 & 0.279 & 0.608 & 0.688 & 0.288 & 0.540 \\
\bottomrule
\end{tabular}%
}
\end{table*}

\section{Results}
As shown in Table \ref{tab:lexical_corr}, path signature embeddings exhibit substantially stronger alignment with lexical string metrics compared to both average pooling and pre-trained deep learning baselines. Most notably, average pooling and Transformer-based models (Minilm, Qwen, Text2Vec) fail almost entirely to capture character-level edit distances, yielding near-zero or very low Pearson correlations with Levenshtein and Phoneme Levenshtein (e.g., Minilm at 0.025 and 0.031; average pooling at 0.036 and 0.017). In contrast, signature embeddings at depth 1 elevate these correlations to 0.367 and 0.361 respectively. This stark difference arises because character-level edits fundamentally alter the sequential trajectory of the word vector path, a geometric property naturally captured by the iterated integrals of the signature, whereas deep models semantically abstract away from surface form. While pooling and deep models maintain reasonable alignment with Word Jaccard and Char TF-IDF by capturing set or subword overlap, depth 2 signatures surpass all baselines on Word Jaccard (0.690 vs. 0.608 for Minilm), demonstrating that incorporating sequential interactions complements and enhances lexical sensitivity.

Analyzing the ablation across Word2Vec dimensions reveals that lexical correlation rapidly saturates. Increasing the word vector dimension from 50 to 400 yields negligible improvements across all configurations, suggesting that coarse lexical geometry is captured in relatively low-dimensional token spaces. Conversely, increasing signature depth profoundly shifts the correlation profile. Depth 1 signatures excel at capturing linear trajectory shifts induced by character edits but over-penalize word reordering, causing Word Jaccard correlation to drop to 0.352. Depth 2 recovers high Word Jaccard alignment (0.690) by encoding second-order interaction areas, effectively balancing strict sequential fidelity with set-based overlap. Depth 3 provides only marginal gains over depth 2 at dimension 50, and computational memory constraints prevent its evaluation at higher dimensions due to the combinatorial explosion of the signature feature space. Finally, aggressive PCA compression highlights extreme redundancy in the raw signature space. Reducing depth 2 signatures to just 10\% of their principal components slightly attenuates Levenshtein alignment (from 0.351 to 0.292) but surprisingly preserves or slightly improves Word Jaccard and Char TF-IDF correlations, indicating that the most interpretable lexical variance is concentrated in a highly compact subspace.

A fundamental distinction between the representation paradigms is their sensitivity to word order. Average pooling is explicitly permutation invariant, mapping any reordering of the same word set to an identical vector and yielding an order sensitivity ratio of zero in our positional probes. Path signatures, by construction, map different permutations of the same word sequence to distinct geometric trajectories. The iterated integrals at depth 2 and beyond explicitly encode the sequence of token intersections, meaning that a shuffled sentence alters the signed area enclosed by the path. Our text positioning study confirms that signature distances correlate positively with the normalized inversion disorder of shuffled sentences, verifying that the embedding space structurally preserves and quantifies positional information rather than collapsing it.

\section{Conclusion}
In this work, we investigated the construction of sentence embeddings using path signatures over Word2Vec trajectories, positioning them as an interpretable and sequentially aware alternative to conventional pooling methods. Our empirical analysis demonstrated that signature embeddings strongly align with lexical string metrics, recovering character-level edit distances that permutation-invariant pooling fundamentally obliterates. We found that depth 2 signatures optimally balance character-level and word-level lexical sensitivity, and that the most critical lexical information resides in an extremely compact subspace, surviving aggressive PCA compression. Furthermore, we verified that these embeddings structurally preserve word order, quantifying positional perturbations rather than collapsing them. While the combinatorial explosion of the signature space limits practical scaling to higher depths, path signatures offer a mathematically rigorous lens into how continuous text representations encode discrete lexical properties. Future work may explore hybrid architectures that integrate the sequential fidelity of signatures with the semantic depth of attention mechanisms.

\end{multicols}
\bibliographystyle{unsrt}
\nocite{*}
\bibliography{references}
\newpage
\appendix

\begin{multicols}{2}

\section{Experimental Configuration Details}
\label{sec:appendix_config}

To ensure full reproducibility, we detail the complete experimental configuration used in our pipeline. A global random seed of 42 is applied across Python, NumPy, and PyTorch to guarantee deterministic execution where possible.

\subsection{Datasets}
\begin{itemize}
    \item \textbf{STS-12} (\texttt{mteb/sts12-sts}): The training split is used solely for Word2Vec corpus augmentation, while the test split is reserved strictly for evaluation. No evaluation leakage occurs.
    \item \textbf{BookCorpus} (\texttt{rojagtap/bookcorpus}): Streamed with a cap of 100,000 sentences to serve as the primary large-scale corpus for Word2Vec training, merged with the deduplicated STS-12 training sentences.
\end{itemize}

\subsection{Word2Vec Training}
We train gensim Word2Vec models from scratch on the combined corpus. We enforce deterministic training by setting workers to 1.
\begin{itemize}
    \item \textbf{Architectures}: Skip-gram.
    \item \textbf{Dimensions}: $\{50, 100, 200, 400\}$.
    \item \textbf{Hyperparameters}: Window size = 5, minimum word count = 5, negative samples = 5, subsampling threshold = $10^{-3}$, epochs = 5.
\end{itemize}

\subsection{Path Signatures and Pooling}
Paths are constructed by connecting sequential word vector waypoints. Constant paths (zero displacement) are filtered out prior to signature computation.
\begin{itemize}
    \item \textbf{Signature Depths}: Configured by dimension to avoid out-of-memory errors due to combinatorial feature explosion: dimensions 50 use depths $\{1, 2, 3\}$; dimensions 100, 200 and 400 use depths $\{1, 2\}$.
    \item \textbf{Representations}: signature are computed via \texttt{iisignature}.
    \item \textbf{Pooling Baselines}: Mean pooling of word vectors are computed as permutation-invariant baselines.
    \item \textbf{Distance Metrics}: Cosine distances are calculated between all pair representations.
\end{itemize}

\subsection{Dimensionality Reduction (PCA)}
Given the high dimensionality of raw signature spaces, we apply PCA in two regimes:
\begin{itemize}
    \item \textbf{Diagnostic PCA}: Capped at 50 components to compute explained variance ratios and intrinsic dimensionality estimates for variance targets of 90\%, 95\%, and 99\%.
    \item \textbf{Distance PCA}: Applied prior to distance computation with retention ratios of $\{50\%, 20\%, 10\%, 1\%\}$ of the raw signature dimension $D$. The number of components is hard-capped at 512 to manage SVD runtime and memory.
\end{itemize}

\subsection{Lexical and String Metrics}
\begin{itemize}
    \item \textbf{Character TF-IDF}: Cosine distance on character and character word-boundary n-grams (range 3 to 5).
    \item \textbf{MinHash}: Jaccard distance estimated with 128 permutations.
    \item \textbf{String Distances}: Levenshtein, Jaro-Winkler, Word Jaccard (on token sets), and Phoneme Levenshtein (Levenshtein on concatenated Metaphone codes).
\end{itemize}

\subsection{Pre-trained Sentence Encoders}
We benchmark against three Transformer-based models using cosine similarity on L2-normalized embeddings:
\begin{itemize}
    \item \textbf{MiniLM}: \texttt{sentence-transformers/all-MiniLM-L6-v2} (batch size 64).
    \item \textbf{Qwen Embed}: \texttt{Qwen/Qwen3-Embedding-0.6B} (batch size 16).
    \item \textbf{Text2Vec}: \texttt{shibing624/text2vec-base-multilingual} (batch size 32).
\end{itemize}

\section{Spearman and Kendall Tau Table}
\label{apd:second}

\begin{table*}[t]
\centering
\caption{Spearman correlation of embedding distances against five lexical string metrics across varying Word2Vec dimensions and signature configurations.}
\label{tab:lexical_corr_spearman}
\vspace{0.1cm}
\resizebox{\textwidth}{!}{%
\begin{tabular}{llccccc}
\toprule
\textbf{Configuration} & \textbf{Dim} & \textbf{Levenshtein} & \textbf{Jaro-Winkler} & \textbf{Word Jaccard} & \textbf{Phoneme Lev.} & \textbf{Char TF-IDF} \\
\midrule
\multirow{4}{*}{Avg Pooling}
    & 50  & 0.057 & 0.511 & 0.753 & 0.076 & 0.639 \\
    & 100 & 0.069 & 0.524 & 0.758 & 0.089 & 0.628 \\
    & 200 & 0.076 & 0.527 & 0.757 & 0.097 & 0.618 \\
    & 400 & 0.077 & 0.527 & 0.756 & 0.098 & 0.619 \\
\midrule
    Sbert Text2Vec & - & 0.292 & 0.523 & 0.533 & 0.280 & 0.456 \\
    Qwen Embed & - & 0.218 & 0.494 & 0.622 & 0.214 & 0.727 \\
    Minilm Embed & - & 0.179 & 0.488 & 0.675 & 0.178 & 0.772 \\
\midrule
\multirow{4}{*}{Sig (depth 1)}
    & 50  & 0.378 & 0.566 & 0.378 & 0.373 & 0.260 \\
    & 100 & 0.388 & 0.586 & 0.382 & 0.382 & 0.253 \\
    & 200 & 0.387 & 0.589 & 0.380 & 0.380 & 0.246 \\
    & 400 & 0.385 & 0.588 & 0.376 & 0.378 & 0.241 \\
\midrule
\multirow{4}{*}{Sig (depth 2)}
    & 50  & 0.404 & 0.605 & 0.674 & 0.403 & 0.511 \\
    & 100 & 0.405 & 0.611 & 0.673 & 0.405 & 0.500 \\
    & 200 & 0.408 & 0.610 & 0.670 & 0.408 & 0.491 \\
    & 400 & 0.410 & 0.611 & 0.668 & 0.410 & 0.481 \\
\midrule
    Sig (depth 3)
    & 50  & 0.419 & 0.618 & 0.663 & 0.419 & 0.482 \\
\midrule
\multirow{4}{*}{Sig (depth 2, PCA 10\%)}
    & 50  & 0.344 & 0.596 & 0.689 & 0.349 & 0.544 \\
    & 100 & 0.335 & 0.607 & 0.694 & 0.341 & 0.542 \\
    & 200 & 0.332 & 0.606 & 0.688 & 0.339 & 0.532 \\
    & 400 & 0.335 & 0.607 & 0.687 & 0.342 & 0.532 \\
\bottomrule
\end{tabular}%
}
\end{table*}

\begin{table*}[t]
\centering
\caption{Kendall $\tau$ correlation of embedding distances against five lexical string metrics across varying Word2Vec dimensions and signature configurations.}
\label{tab:lexical_corr_kendall}
\vspace{0.1cm}
\resizebox{\textwidth}{!}{%
\begin{tabular}{llccccc}
\toprule
\textbf{Configuration} & \textbf{Dim} & \textbf{Levenshtein} & \textbf{Jaro-Winkler} & \textbf{Word Jaccard} & \textbf{Phoneme Lev.} & \textbf{Char TF-IDF}\\
\midrule
\multirow{4}{*}{Avg Pooling}
    & 50  & 0.035 & 0.363 & 0.572 & 0.048 & 0.560 \\
    & 100 & 0.043 & 0.372 & 0.579 & 0.057 & 0.567 \\
    & 200 & 0.048 & 0.375 & 0.578 & 0.062 & 0.567 \\
    & 400 & 0.048 & 0.374 & 0.577 & 0.063 & 0.566 \\
\midrule
    Sbert Text2Vec & - & 0.201 & 0.371 & 0.377 & 0.193 & 0.368 \\
    Qwen Embed & - & 0.153 & 0.351 & 0.444 & 0.149 & 0.431 \\
    Minilm Embed & - & 0.123 & 0.349 & 0.496 & 0.122 & 0.480 \\
\midrule
\multirow{4}{*}{Sig (depth 1)}
    & 50  & 0.262 & 0.390 & 0.260 & 0.258 & 0.265 \\
    & 100 & 0.270 & 0.407 & 0.264 & 0.266 & 0.269 \\
    & 200 & 0.269 & 0.409 & 0.262 & 0.265 & 0.267 \\
    & 400 & 0.266 & 0.407 & 0.258 & 0.262 & 0.263 \\
\midrule
\multirow{4}{*}{Sig (depth 2)}
    & 50  & 0.279 & 0.433 & 0.499 & 0.280 & 0.494 \\
    & 100 & 0.280 & 0.438 & 0.497 & 0.281 & 0.492 \\
    & 200 & 0.282 & 0.438 & 0.494 & 0.283 & 0.490 \\
    & 400 & 0.283 & 0.438 & 0.492 & 0.284 & 0.488 \\
\midrule
    Sig (depth 3)
    & 50  & 0.289 & 0.443 & 0.489 & 0.290 & 0.485 \\
\midrule
\multirow{4}{*}{Sig (depth 2, PCA 10\%)}
    & 50  & 0.236 & 0.425 & 0.511 & 0.241 & 0.503 \\
    & 100 & 0.229 & 0.435 & 0.514 & 0.235 & 0.506 \\
    & 200 & 0.228 & 0.434 & 0.508 & 0.234 & 0.501 \\
    & 400 & 0.230 & 0.435 & 0.507 & 0.236 & 0.500 \\
\bottomrule
\end{tabular}%
}
\end{table*}

\end{multicols}
\end{document}